# Sélection de la structure d'un perceptron multicouches pour la réduction d'un modèle de simulation d'une scierie.

Philippe THOMAS, Andre THOMAS

Centre de Recherche en Automatique de Nancy CRAN-UMR 7039), Nancy-Université, CNRS

ENSTIB 27 rue du merle blanc, B.P. 1041

88051 Epinal cedex 9 France

philippe.thomas@cran.uhp-nancy.fr

**Résumé— La simulation est souvent utilisée pour évaluer la pertinence d'un Programme Directeur de Production (PDP) ou pour en évaluer l'impact sur des scénarii d'ordonnancement détaillés de la fabrication. Dans ce cadre, nous nous proposons de réduire la complexité d'un modèle de simulation en exploitant un perceptron multicouches. Une phase primordiale lors de la modélisation d'un système à l'aide d'un perceptron multicouches reste la détermination de la structure du réseau. Nous proposons de comparer et d'utiliser divers algorithmes d'élagage pour déterminer la structure optimale du perceptron multicouches utilisé pour réduire la complexité du modèle de simulation de notre cas d'application : une scierie.**

**Mots clés— perceptron multicouches, réseau de neurones, élagage, sélection de structure, modèle réduit, simulation, re-ordonnancement.**

## I. INTRODUCTION

La simulation du fonctionnement d'un atelier est utile pour l'évaluation et/ou la validation d'un ordonnancement prévisionnel [23], ou d'un re-ordonnancement des tâches suite à une perturbation significative. Dans une philosophie de gestion de type « management par les contraintes », l'optimisation du fonctionnement d'une chaîne de fabrication passe par la maximisation du taux de charge des goulots. Celui-ci est le critère principal pour évaluer un PDP. Pour ces raisons, il est intéressant d'utiliser la simulation dynamique des flux [34].

Actuellement, les simulations sont effectuées avec des modèles de plus en plus complexes, ceci grâce au développement des moyens de calcul informatique. Cependant, les modèles complexes obligent leurs concepteurs à faire face à des problèmes pour lesquels ils n'ont pas été formés, et que Page *et al.* [28] appellent « Problems of Scale ». Aussi, de nombreux auteurs ont rappelé l'importance d'utiliser des modèles de simulation simples [2 ; 7 ; 26 ; 29 ; 42].

Par ailleurs, les réseaux de neurones, grâce à leur capacité d'être des « approximateurs universels », ont prouvé leur capacité à extraire de données d'expérimentation des modèles performants, sans avoir à effectuer d'hypothèses sur la forme générale de ces derniers [38]. Aussi, ils commencent à faire leur apparition dans le cadre général de la supply chain [6 ; 31].

Des premiers travaux [39 ; 40] ont conduit à la proposition d'une approche de réduction de modèle de simulation par réseaux de neurones. Cependant, comme pour toute application neuronale, un point crucial de la phase de modélisation reste la détermination de la structure. En effet, même si les travaux de Cybenko [10] et Funahashi [15] ont montré qu'une seule couche cachée utilisant des fonctions d'activation du type sigmoïdal était suffisant pour pouvoir approximer toute fonction non linéaire avec la précision voulue, rien n'est dit a priori sur le nombre de neurones cachés à utiliser.

Pour répondre à ce problème, diverses techniques ont été proposées comme les méthodes de régularisation telle que l'arrêt prématuré (early stopping) [13] ou les méthodes de pénalisation [1 ; 4]. D'autres auteurs ont proposés de construire de manière itérative la couche cachée [5 ; 24 ; 30 ; 32]. La dernière approche consiste à exploiter une structure incluant un grand nombre de neurones cachés puis d'éliminer ces derniers en commençant par les moins significatifs [9 ; 18 ; 22 ; 27 ; 37 ; 43 ; 45].

D'autre part, comme pour tout problème de modélisation, la sélection des variables d'entrée est une tâche primordiale et il est indispensable que l'ensemble des variables d'entrée soit aussi réduit que possible afin de limiter le nombre de paramètres du modèle, toujours dans le but d'éviter les phénomènes de surapprentissage. Plusieurs techniques de sélection de variables basées sur les statistiques notamment ont été proposées. On peut citer, par exemple, l'analyse en composante principale [20], l'analyse en composante curviligne [11] ou encore la méthode du descripteur sonde [12]. D'autres méthodes ont été proposées pour sélectionner les variables dans le cadre strict des réseaux de neurones [3 ; 8 ; 16 ; 21]. Peu de ces méthodes permettent de simultanément sélectionner les variables et d'éliminer les paramètres superflus. Nous pouvons citer entre autres [14 ; 18 ; 33].

Dans ce cadre, nous avons proposé un nouvel algorithme de sélection de structure dérivé de celui proposé par Engelbrecht [14] dont nous avons comparé les performances avec les algorithmes Engel [14] et N2PFA [33]. Ce travail est présenté à CIFA'08 [41].

L'objectif de ce travail est d'exploiter cet algorithme sur un cas réel en concordance avec l'algorithme N2PFA pour déterminer la structure optimale du réseau de neurones qui doit être utilisée dans le modèle réduit de simulation d'une scierie.

Nous allons maintenant présenter la méthode de réduction de modèle utilisée [39] et nous rappellerons très succinctement la structure et les notations du perceptron multicouches utilisées. Nous poursuivrons en présentant le cas d'application qui est une scierie spécialisée dans la découpe de conifères ainsi que les modèles de simulation, complet et réduit. Nous finirons en présentant l'approche utilisée pour déterminer la structure optimale du perceptron multicouches utilisé dans le modèle réduit. Nous en profiterons pour comparer les performances de l'algorithme proposé [41] avec ceux proposés par Engelbrecht [14] et Setiono [33] sur notre cas réel.

## II. LA REDUCTION DE MODELE

### A. L'algorithme de réduction de modèle [39]

La problématique de la réduction de modèle a été initiée par Zeigler [44] pour qui la complexité d'un modèle est relative au nombre d'éléments, de connexions et de calculs du modèle.

Les premières techniques de simplification en modélisation ont été proposées par Innis et Rexstad [19]. Depuis lors, un certain nombre d'auteurs se sont intéressés à ce problème dans les cadres particuliers. L'algorithme de réduction utilisé ici [39] est une extension de celui proposé par Thomas et Charpentier [35] :

1.  Identifier le goulet structurel (Poste de charge (PdC) qui, depuis plusieurs années, est majoritairement contrainte de capacité).
2.  Pour la liasse d'Ordre de Fabrication (OF) du PDP considéré, identifier le goulet conjoncturel.
3.  Parmi les autres PdC, identifier ceux (postes de synchronisation) satisfaisant aux deux conditions :
    -   utilisé conjointement à un poste goulet par au moins un OF,
    -   largement utilisé lorsque l'ensemble des OF est considéré.
4.  Si tous les OF ont été considérés, passer à 5. sinon retourner en 3.
5.  Modéliser par un réseau de neurones l'intervalle situé entre chacun des PdC précédemment trouvés.

### B. Le perceptron multicouches

Nous allons ici rappeler la structure du perceptron multicouches utilisé. Le réseau présenté par la figure 1 est composé de neurones interconnectés en trois couches successives.

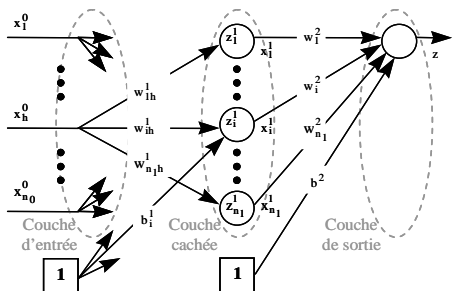

Figure 1. L'architecture du perceptron multicouches

La première couche est composée de neurones « transparents » qui n'effectuent aucun calcul mais simplement distribuent leurs entrées à tous les neurones de la couche suivante appelée couche cachée. Les neurones de la couche cachée (figure 1), dont un exemple peut être représenté par la figure 2, reçois les $n_0$ entrées $\{x_1^0, \cdots, x_{n_0}^0\}$ de la couche d'entrée avec les poids associés $\{w_{i1}^0, \cdots, w_{in_0}^0\}$. Ce neurone commence par calculer la somme pondérée de ses $n_0$ entrées :

$$z_i^1 = \sum_{h=1}^{n_0} w_{ih}^1 . x_h^0 + b_i^1 \qquad (1)$$

où $b_i^1$ est un biais (ou seuil).

La sortie du neurone caché est obtenue en transformant la somme (1) par l'intermédiaire de la fonction d'activation g(.) :

$$x_i^1 = g(z_i^1). \qquad (2)$$

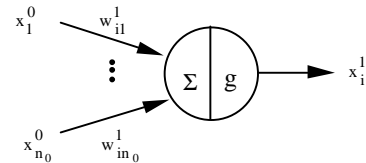

Figure 2. Le neurone i de la couche cachée

Bien que de nombreuses fonctions d'activations ait été proposées, la fonction g(.) est généralement la tangente hyperbolique [38] :

$$g(x) = \frac{2}{1 + e^{-2x}} - 1 = \frac{1 - e^{-2x}}{1 + e^{-2x}}. \qquad (3)$$

Le neurone de la dernière couche (ou couche de sortie) utilise une fonction d'activation linéaire et n'effectue donc qu'une simple somme pondérée de ses entrées :

$$z = \sum_{i=1}^{n_1} w_i^2 . x_i^1 + b \qquad (4)$$

où $w_i^2$ sont les poids connectant les sorties des neurones cachés au neurone de sortie et b est le biais du neurone de sortie.

## III. LE CAS D'ETUDE

### A. La scierie

Le cas d'application étudié est une scierie, appartenant au premier scieur français, située dans le massif des Vosges en France et spécialisée dans la transformation de conifères en planches.

En 2001, cette scierie avait une capacité de 270 000 m³/an. Elle générait un chiffre d'affaire de 53 millions d'euros et employait 300 personnes.

La ligne de sciage peut être décomposée en trois parties principales. Afin d'en comprendre le fonctionnement, nous allons suivre le cheminement d'une grume de son entrée dans la chaîne, à sa sortie sous forme de diverses planches.

La première partie de la ligne de sciage est constituée de la ligne canter présentée à la figure 3. La grume que l'on doit traiter entre dans la chaîne par la droite par l'intermédiaire des convoyeurs RQM1 RQM2 et RQM3. En fonction de ses caractéristiques (scanner MS), notre grume va être dirigée soit vers RQM4 soit vers RQM5 qui font office de stocks. Elle va ensuite être amenée sur la première machine canter puis sur la délignueuse CSMK qui a pour but de transformer notre grume en un parallélépipède rectangle, l'équarri. Cette première étape, qui forme les deux premiers cotés de l'équarri, produit 2 planches (appelées produits secondaires) qui seront expulsées par l'intermédiaire des convoyeurs BT4 et BT5. Le reste de la grume va être éjecté sur RQM6, va subir une rotation de 90° pour être enfin stocké sur RQM7 en attendant de repasser sur la même machine pour faire les deux autres cotés du parallélépipède. A l'issue du deuxième passage, outre l'équarri, on obtient deux autres produits secondaires qui sont également évacués par les convoyeurs BT4 et BT5 en direction de la Tronçonneuse Déligneuse dont l'élément principal est la scie Kockum. L'équarri est ensuite déligné sur la déligneuse MKV en trois planches supplémentaires que nous appellerons produits principaux. Ces produits principaux sont directement dirigés vers l'ébouteur.

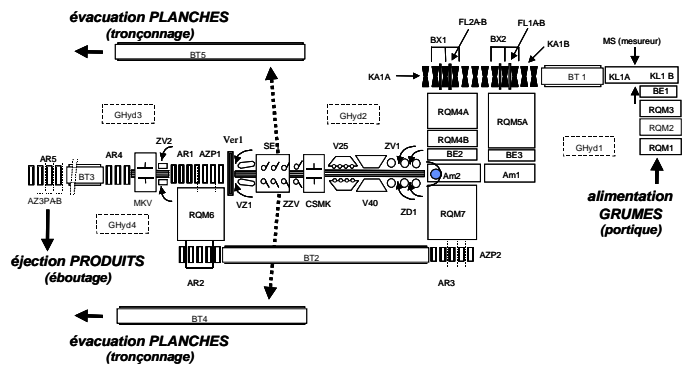

Figure 3. La ligne canter

La figure 4 présente la deuxième partie de la ligne de sciage qui correspond au tronçonnage délignage et dont la machine principale est la scie Kockums.

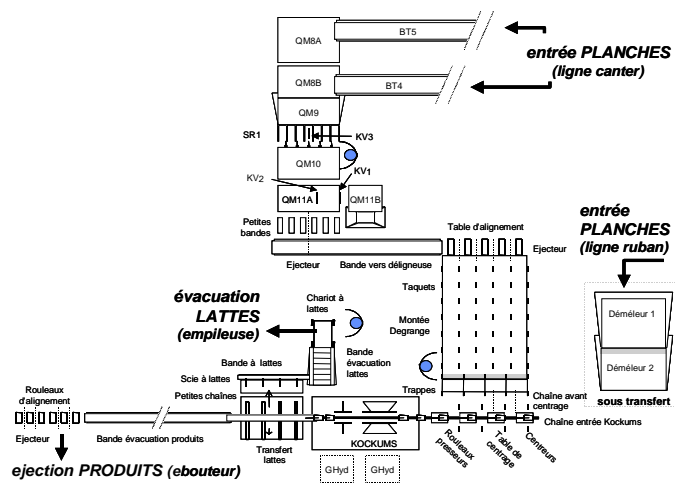

Figure 4. La tronçonneuse-deligneuse (Kockums)

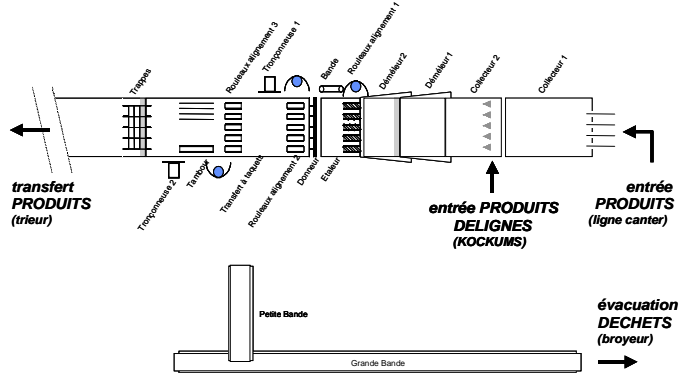

Figure 5. L'ébouteur

En entrée de l'ébouteur se trouvent deux collecteurs (collecteur 1 et collecteur 2) qui permettent l'admission sur la ligne respectivement des produits secondaires en provenance de la scie kockums, et des produits principaux en provenance directe de la ligne canter. Les pièces sont démêlées et mises sur un tapis à taquets. La tronçonneuse 1 va permettre d'effectuer une coupe de purge des défauts tandis que la tronçonneuse 2 permet de donner sa longueur finale au produit. Des travaux précédents [34] ont montré ce c'est cette machine, l'ébouteur, qui est le poste goulot de la chaîne de sciage.

### B. Les modèles de simulation

Des travaux précédents [34] ont permis de construire le modèle complet de la ligne de sciage. Ce modèle est présenté à la figure 6.

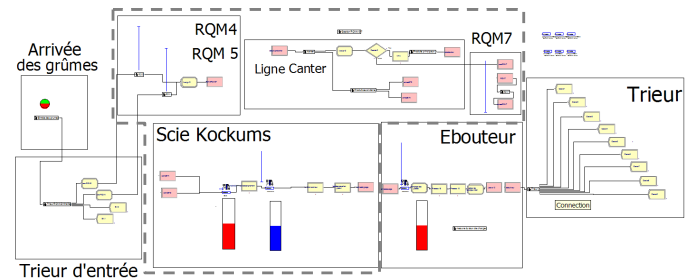

Figure 6. Le modèle complet

Ce modèle est suffisamment complexe pour pouvoir simuler assez précisément le comportement de la ligne de sciage précédemment décrite.

Cependant, nous savons que le poste goulot d'étranglement de cette ligne de sciage est l'ébouteur [34 ; 39]. Donc, un pilotage optimal de la ligne nécessite d'optimiser la charge du poste goulot, en accord avec le principe donné par la méthode O.P.T. que « toute heure perdue sur un poste goulot est une heure perdue pour le système. » [17 ; 25].

Dans ce cadre, il semble bien que la modélisation des stocks RQM4 et 5, de la ligne Canter, du stock RQM7, de la gestion des RQM4-5 et 7 et même de la scie kockums soit superflue. Donc toute la partie entourée par des tirets gris sur la figure 6 n'apporte pas d'information directe et utile pour l'évaluation d'un PDP ou la mise en place d'un re-ordonnancement pour lesquels nous aurions des indicateurs de performance tels que délai global ou taux de charge du goulot... En effet, seul les instants d'arrivée des produits en entrée de l'ébouteur vont nous permettre de simuler la charge de ce dernier. C'est pourquoi, nous proposons de remplacer toute la partie entourée de tiret gris sur la figure 6 par un perceptron multicouches qui va transformer les informations en entrée du modèle et fournies par les modules « arrivée des grumes » et « trieur d'entrée » en des instants d'arrivée de produits en entrée de l'ébouteur sans que les chemins suivis ou les transformations subies soient apparentes.

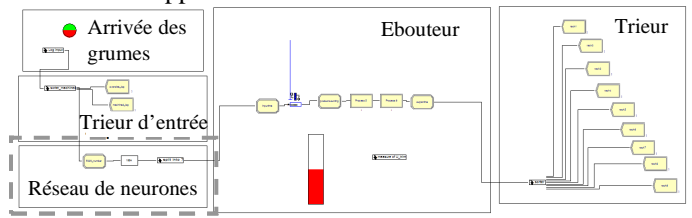

Figure 7. Le modèle réduit

Nous obtenons ainsi un modèle réduit (figure 7) où toute une partie du système est modélisée à l'aide d'un perceptron multicouches dont il nous reste à déterminer la structure.

## IV. LES RESULTATS

### A. Première approche

Des premiers travaux [39 ; 40] ont permis de sélectionner et collecter les données disponibles en entrée du système. Ces données sont des informations intrinsèques à la grume (longueur, 3 valeurs de diamètres), à son traitement (passage sur RQM4 ou RQM5), à l'état de la ligne (taille du stock d'entrée de l'ébouteur, taux d'utilisation de l'ébouteur, nombre de grumes présentes dans le système), et au plan de coupe (instant et lieu de la découpe du produit, type de produit).

Nous disposons donc d'un jeu de dix variables d'entrées (longueur ; diamGrosBout ; diamMoyen ; diamPetitBout ; produit ; type_piece ; Q_ebouteur ; taux_ebouteur ; Q_RQM ; RQM) que l'on va pouvoir associer aux 12775 pièces produites. Ces dix variables vont constituer les dix entrées du perceptron multicouches.

Ces 12275 pièces vont mettre un certain temps entre le moment où elles entrent dans le système sous forme de grume et le moment où elles entrent dans le stock d'entrée de l'ébouteur. C'est ce temps $\Delta T$ que nous cherchons à déterminer. C'est donc lui qui constituera la sortie de notre réseau de neurones.

Des travaux précédents [40] utilisant l'algorithme OBS [18] avaient montré que 24 neurones cachés son suffisants pour modéliser ce système. Aussi, nous avons choisi d'utiliser 25 neurones cachés dans la structure du modèle initial.

La détermination des paramètres du réseau de neurones s'effectuant par apprentissage supervisé, il est nécessaire de diviser la base de données précédemment constituée en deux jeux. Le premier, appelé jeu d'apprentissage, va nous servir à effectuer l'apprentissage proprement dit. Le deuxième, appelé jeu de validation, nous servira à vérifier que l'apprentissage s'est correctement passé, en particulier, à évaluer la présence ou l'absence de sur-apprentissage.

L'algorithme d'apprentissage utilisé est l'algorithme de Levenberg Marquardt utilisant un critère robuste [38] sur 50000 itérations maximum.

### B. Comparaison des algorithmes

Afin de prendre en compte le fait que l'algorithme d'apprentissage effectue une recherche locale du minimum, 50 jeux de paramètres initiaux ont été obtenus en utilisant une modification de l'algorithme de Nguyen et Widrow [36]. Les 3 algorithmes testés ont tous fonctionné sur les mêmes jeux de paramètres initiaux. Ces algorithmes sont N2PFA [33], Engel [14] et Engel_mod [41].

L'ensemble des résultats obtenus sur ce premier système avec les trois algorithmes testés sont regroupés et synthétisés dans le tableau 1.

Table 1 : résultats des trois algorithmes

la première ligne indique le nombre d'entrées conservées par les algorithmes, tandis que la deuxième donne le nombre de neurones cachés conservés et la troisième fournit le nombre de paramètres (poids et biais) constituant le modèle résultant. Les lignes quatre et cinq fournissent respectivement les valeurs de la somme des carrés des erreurs (NSSE) obtenus pour les 50 jeux de données d'apprentissage et de validation respectivement. La dernière ligne indique le temps mis par l'algorithme pour se dérouler.

Pour chacune des lignes, les valeurs minimale maximale et moyenne du paramètre considéré sont relevées pour l'algorithme considéré pour les 50 jeux de paramètres initiaux. Nous trouvons également deux pourcentages. Le premier indique le pourcentage de jeux de poids initiaux fournissant une valeur inférieure à la valeur moyenne, et le deuxième, le pourcentage de jeux de poids initiaux fournissant une valeur supérieure à la valeur moyenne. Chacune des colonnes correspond à chacun des algorithmes testés.

Tout d'abord, si l'on étudie le nombre d'entrées conservées dans les modèles, on constate que les trois algorithmes ont des comportements très différents. En effet, si Engel_mod conserve toujours toutes les entrées, Engel va au contraire dans certains cas jusqu'à supprimer toutes les entrées. N2PFA a lui un comportement beaucoup plus réaliste dans son élimination des variables d'entrée.

En ce qui concerne le nombre de neurones cachés, on pouvait s'attendre, vu les précédents travaux exploitant l'algorithme OBS que la grande majorité de ces derniers doivent être conservés. Or, si l'algorithme Engel va jusqu'à supprimer tous les neurones cachés dans certains cas, les trois algorithmes convergent vers un nombre optimal de neurones cachés de l'ordre de 2. Cependant, si les algorithmes Engel et Engel_mod donnent quasiment toujours une bonne approximation du nombre de neurones cachés utiles (à l'exclusion des cas aberrants de Engel : 0 entrées ou 0 neurones cachés) avec une moyenne de 2,8 un min de 2 et un max de 5 pour Engel_mod, l'algorithme N2PFA ne parviens aux 2 neurones cachés que dans 6% des cas et dans 28% des cas, il ne supprime même aucun neurone. En ce qui concerne le nombre de paramètres conservés dans le modèle, On constate que l'algorithme Engel_mod est celui qui en conserve le moins (les résultats de Engel sont biaisés par les cas aberrants). Ceci est en particulier dû au fait qu'il conserve un optimum de neurones cachés, mais de surcroît, cet algorithme supprimant les paramètres un par un, même s'il ne parvient pas à supprimer une entrée complètement, il supprime tout de même un certain nombre de poids connectant les entrées aux neurones cachés.

Les valeurs de NSSE pour les jeux d'apprentissage et de validation permettent de confirmer ou infirmer un choix de structure. Ces valeurs sont difficilement comparables entre algorithmes puisque N2PFA effectue un réapprentissage sur 50 itérations après chaque élimination d'entrée ou de neurone caché ce qui n'est pas le cas pour les deux autres algorithmes. On peut noter cependant que pour les trois algorithmes ce sont les structures conservant uniquement 2 neurones cachés qui fournissent les meilleurs résultats de NSSE ce qui confirme un choix de structure conservant uniquement 2 neurones cachés.

Enfin, la dernière ligne fournit les temps (en s) de calcul mis par les trois algorithmes. Si les deux algorithmes Engel et Engel_mod utilisent des temps relativement similaires, le déroulement de l'algorithme N2PFA va lui durer, en moyenne,

| | | engel | | | engel_mod | | | N2PFA | | |
|---|---|---|---|---|---|---|---|---|---|---|
| | | min | moy | max | min | moy | max | min | moy | max |
| Nb_I | val | 0 | 8,14 | 10 | 10 | 10 | 10 | 5 | 8,62 | 10 |
| | % | 62% | < > | 38% | | < > | | 38% | < > | 62% |
| Nb_H | val | 0 | 2,26 | 5 | 2 | 2,8 | 5 | 2 | 18,82 | 25 |
| | % | 72% | < > | 28% | 48% | < > | 52% | 40% | < > | 60% |
| Nb_θ | val | 1 | 25,08 | 61 | 24 | 34,2 | 61 | 21 | 202,0 | 301 |
| | % | 72% | < > | 28% | 48% | < > | 52% | 68% | < > | 58% |
| NSSE_ID | val | 268250 | 381138 | 538740 | 264590 | 367401 | 509920 | 154610 | 248417 | 424620 |
| | % | 58% | < > | 42% | 58% | < > | 42% | 64% | < > | 36% |
| NSSE_val | val | 265350 | 407081 | 639370 | 285100 | 393779 | 575600 | 175810 | 266351 | 502580 |
| | % | 58% | < > | 42% | 58% | < > | 42% | 68% | < > | 32% |
| temps | val | 12,45 | 104,51 | 519,31 | 53,53 | 150,58 | 512,42 | 167,74 | 457,74 | 1570,6 |
| | % | 64% | < > | 36% | 76% | < > | 24% | 56% | < > | 44% |

trois fois plus longtemps pour durer près d'une demi heure dans le pire des cas.

En résumé, l'algorithme Engel conduit dans 18% des cas à une structure aberrante sans entrée ou sans neurone caché. L'algorithme N2PFA est le seul qui parvient à effectuer une sélection parmi les variables d'entrées. Cependant il ne parvient que rarement à déterminer correctement le nombre de neurones cachés à conserver. Dans seulement 14% des cas, il conserve moins de 6 neurones cachés. De surcroît, cet algorithme prend près de trois fois plus de temps pour se dérouler. Enfin, si l'algorithme Engel_mod ne parvient pas à supprimer la moindre entrée, il estime correctement le nombre de neurones cachés à inclure dans la structure et ce dans un temps raisonnable.

## C. Association N2PFA Engel_mod

Considérant Les résultats précédents, nous avons choisi d'associer les algorithmes Engel_mod et N2PFA pour déterminer la structure optimale du réseau. Pour ce faire nous proposons de commencer par appliquer l'algorithme Engel_mod sur une structure initiale, puis d'appliquer l'algorithme N2PFA sur la structure trouvée. L'objectif est d'optimiser le temps de calcul. En effet l'algorithme Engel_mod est le plus rapide des deux et il permet d'éliminer plus de paramètres que son concurrent. Dans un deuxième temps, l'algorithme N2PFA travaillant sur une structure plus petite, devrait prendre moins de temps.

| | | engel_mod + N2PFA | | |
| | | min | moy | max |
|---|---|---|---|---|
| Nb_I | val | 5 | 7,98 | 10 |
| | % | 62% | < > | 38% |
| Nb_H | val | 1 | 2,16 | 3 |
| | % | 72% | < > | 28% |
| Nb_θ | val | 10 | 15,42 | 25 |
| | % | 72% | < > | 28% |
| NSSE_ID | val | 179070 | 232323 | 463450 |
| | % | 80% | < > | 20% |
| NSSE_val | val | 189300 | 244752 | 487760 |
| | % | 80% | < > | 20% |
| moyenne erreur_ID | val | 2,28E-11 | 5,94 | 272,69 |
| | % | 98% | < > | 0,02 |
| écart type erreur_ID | val | 423,2 | 478,5974 | 680,82 |
| | % | 80% | < > | 20% |
| moyenne erreur_val | val | 0,02 | 14,22 | 270,79 |
| | % | 84% | < > | 16% |
| écart type erreur_val | val | 435,06 | 490,69 | 698,31 |
| | % | 80% | < > | 20% |
| temps | val | 502,58 | 629,18 | 1189,5 |
| | % | 68% | < > | 32% |

Table 2 : résultats de l'association

Le tableau 2 regroupe l'ensemble des résultats concernant les 50 essais. Tout d'abord, nous pouvons constater que le temps de calcul a bien été fortement diminué. L'ensemble de l'élagage prend tout de même plus de 10 min en moyenne.

On constate également que dans la quasi-totalité des cas, nous retrouvons 2 neurones dans la couche cachée. En ce qui concerne les entrées 8 sont conservées en moyenne.

Afin de sélectionner la meilleure structure, nous allons privilégier celle qui nous donne les plus petites erreurs moyennes sur les jeux d'apprentissage et de validation tout en fournissant de bons résultats sur les NSSE, et donc sur les écart-types. La structure ainsi sélectionnée comprend 8 entrées, 2 neurones cachés et 21 paramètres. Les erreurs moyennes relevées en apprentissage $(1,9.10^{-9})$ et en validation (0,018) sont très faibles. De même les écarts type de l'erreur sont parmi les plus faibles (437,56 en apprentissage et 456,17 en validation).

Nous pouvons constater que cette structure est celle vers laquelle converge la majorité des essais. La figure 8 présente la structure sélectionnée. Nous pouvons constater que sur les dix entrées initialement sélectionnées, seul huit sont conservées. En particulier, les variables longueur de la grume et type de produit n'apportent pas d'informations significatives. Pour la variable produit, cet élagage pouvait être prévu puisque la variable type_pièce intègre naturellement cette information. Ces deux variables sont donc fortement corrélées et la variable type_pièce est plus informative.

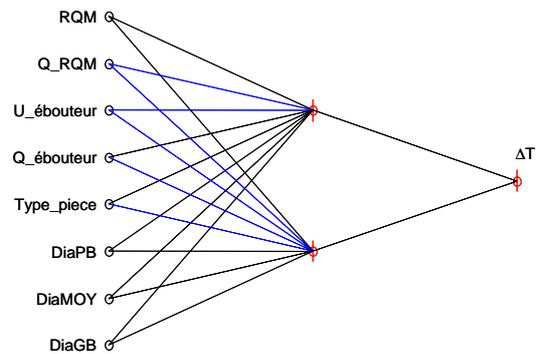

Figure 8. Structure du réseau

Nous pouvons cependant constater que les écarts types des erreurs restent important (et ce, quel que soit le nombre de neurones cachés utilisés dans le réseau). Ceci peut s'expliquer par le fait qu'un certain nombre de variables explicatives du système ne sont pas utilisées par le réseau de neurones. Par exemple, on peut citer, le pourcentage de grumes mitraillées, la répartition des grumes sur les différents RQM…

## V. CONCLUSION

Dans cet article, nous avons comparé trois algorithmes d'élagage de la structure d'un perceptron multicouches sur un cas d'application réel. Les résultats montrent bien l'intérêt des algorithmes comparés et surtout leur complémentarité. En effet, l'utilisation conjointe de deux de ces algorithmes permet de diminuer le temps de calcul tout en améliorant les résultats. Dans nos travaux futurs, nous allons exploiter cette approche de réduction de modèles de simulation dans le cadre des systèmes contrôlés par le produit.